\documentclass{bmvc2k}

\usepackage{wrapfig}
\usepackage{amssymb}
\usepackage{multirow}
\usepackage{adjustbox}
\usepackage{float}
\usepackage{graphicx}
\usepackage{array}
\newcolumntype{P}[1]{>{\centering\arraybackslash}p{#1}}
\newcolumntype{M}[1]{>{\centering\arraybackslash}m{#1}}

\title{Deep Textured 3D Reconstruction of \\Human Bodies}

\addauthor{Abbhinav Venkat}{abbhinav.venkat@research.iiit.ac.in}{1}
\addauthor{Sai Sagar Jinka}{jinka.sagar@reseach.iiit.ac.in}{1}
\addauthor{Avinash Sharma}{asharma@iiit.ac.in}{1}

\addinstitution{Center for Visual Information \\ 
Technology (CVIT), \\
Kohli Center for Intelligent Systems \\
(KCIS), \\
IIIT Hyderabad, \\
India}

\runninghead{Abbhinav Venkat, Sai Sagar, Avinash Sharma}{3D Human Reconstruction}

\begin{document}

\maketitle

\begin{abstract}
%Rewrite and reduce
Recovering textured 3D models of non-rigid human body shapes is challenging due to self-occlusions caused by complex body poses and shapes, clothing obstructions, lack of surface texture, background clutter, sparse set of cameras with non-overlapping fields of view, etc. Further, a calibration-free environment adds additional complexity to both - reconstruction and texture recovery.  In this paper, we propose a deep learning based solution for textured 3D reconstruction of human body shapes from a single view RGB image. This is achieved by first recovering the volumetric grid of the non-rigid human body given a single view RGB image followed by orthographic texture view synthesis using the respective depth projection of the reconstructed (volumetric) shape and input RGB image. We propose to co-learn the depth information readily available with affordable RGBD sensors (e.g., Kinect) while showing multiple views of the same object during the training phase.  We show superior reconstruction performance in terms of quantitative and qualitative results, on both, publicly available datasets (by simulating the depth channel with virtual Kinect) as well as real RGBD data collected with our calibrated multi Kinect setup.  

\end{abstract}

%-------------------------------------------------------------------------
\section{Introduction }
\label{sec:intro}
\begin{figure*}[h!]
\begin{center}
\subfigure[]{\includegraphics[width=2cm, height=2.8cm]{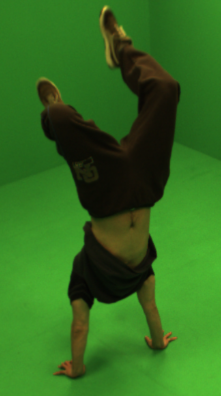} \label{subfig:proba}} 
\subfigure[]{\includegraphics[width=2cm, height=2.8cm]{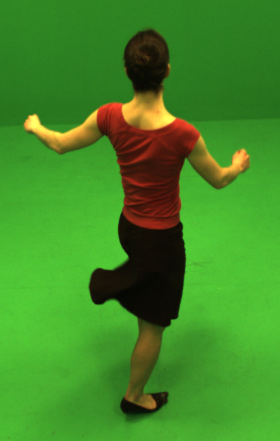} \label{subfig:probb}} 
\subfigure[]{\includegraphics[width=2cm, height=2.8cm]{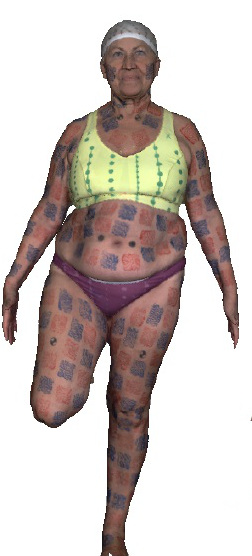} \label{subfig:probc}} 
\subfigure[]{\includegraphics[width=2cm, height=2.8cm]{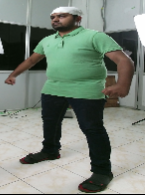} \label{subfig:probd}} 
\caption{Challenges in non-rigid reconstruction. (a) Complex poses (b) Clothing obstructions (c) Shape variations (d) Background clutter.}
\label{fig:problems}
\end{center}
\end{figure*}

Recovering the textured 3D model of non-rigid human shapes from images is of high practical importance in the entertainment industry, e-commerce, health care (physiotherapy), mobile based AR/VR platforms, etc. This is a challenging task as the object geometry of non-rigid human shapes evolve over time, yielding a large space of complex body poses as well as shape variations. In addition to this, there are several other challenges such as self-occlusions by body parts, obstructions due to free form clothing, background clutter (in a non-studio setup), sparse set of cameras with non-overlapping fields of views, sensor noise, etc. Figure~\ref{fig:problems} illustrates some of these challenges. Model based reconstruction techniques attempt to overcome some of these limitations, however, at the cost of loss of accurate geometrical information over the shape surface and are typically applicable only for tight clothing scenarios~\cite{SCAPETOG2005,SMPL2016,dfaust:CVPR:2017}. %Moreover, unlike rigid objects, the object geometry evolves over time in the case of non-rigid objects, hence, learning a model becomes more difficult. Further, this evolving geometry induces non-rigid deformations due to clothing (as seen in Figure~\ref{subfig:probb}), hence resulting in additional challenges when attempting to reconstruct as well as recover the texture.     

Traditionally, calibrated multi-camera setups have been employed to recover textured 3D models through triangulation or voxel carving~\cite{VlasicMIT2008,dfaust:CVPR:2017}. However, these techniques yield  reconstructions with severe topological noise~\cite{SharmaSpectral2011}. Recent attempts replace/augment the capture setup with high resolution depth sensors~\cite{DepthRegis:ToG2014,Fusion4D:2016}, thus making the setup more accurate but less affordable. Nevertheless, the  fundamental limitation of these techniques is the  requirement of a calibrated multi-camera/sensors that restricts their applicability to studio environments. We aim at achieving 3D reconstruction of textured human body models using single/multiple affordable RGB/D sensor(s) in a calibration-free setup. 
%
%Model based reconstruction techniques attempt to overcome some of these limitations, however, at the cost of loss of accurate geometrical information over the shape surface~\cite{SCAPETOG2005,SMPL2016,dfaust:CVPR:2017}. Another approach involves performing non-rigid registration over point clouds (coming from RGBD sensors) to recover textured surface reconstructions~\cite{DepthRegis:ToG2014,Fusion4D:2016}. Nevertheless, even such methods suffer from challenges shown in Figure~\ref{fig:problems}. More importantly, majority of approaches belonging to the three shape reconstruction paradigms listed above expect a high-resolution calibrated multi-camera setup which is expensive and limited in their ubiquitousness. In a calibration-free setting such as ours, motion capture of dynamic scenes is difficult, thus, making both reconstruction and texture recovery not trivial.  
\subsection{Related Work }
\label{sec:related}
Recent advancement in deep networks has enabled learning a class specific 3D structure of a set of objects (e.g. cars, chairs, rooms) using large scale datasets of synthetic 3D models for training the deep networks~\cite{3DShapeNet2015,3DGAN2016,3DR2N22016,PTNet2016,RayMalik2017}. ShapeNet\cite{3DShapeNet2015} proposed a deep network representation with a convolutional deep belief network to give a probabilistic representation of the voxel grid. Along similar lines, 3D Generative Adversarial Networks (GAN's) were proposed to learn a probabilistic latent space of rigid objects (such as chairs, tables) in~\cite{3DGAN2016}. \cite{PTNet2016} proposed an encoder-decoder network that utilizes observations from the 2D space, and without supervision from 3D, performs reconstruction for a few classes of objects. This relationship between 2D and 3D is further exploited in~\cite{RayMalik2017} where they define a loss function in terms of ray consistency and train for single view reconstruction. 
However, these methods have been employed only for rigid object reconstruction. 

%There are only few early efforts in literature around non-rigid object reconstruction using deep learning, which are not directly extendable to non-rigid human body shapes. 
In regard to non-rigid reconstruction, ~\cite{SurfNet2017} proposed to directly achieve the surface reconstruction by learning a mapping between 3D shapes and the geometry image representation (introduced in ~\cite{GeometryImageSinha2016}). One of the key limitations of their method is that it is only suitable for genus-0 meshes. This constraint is frequently violated in real-world human body shapes due to topological noise~\cite{SharmaSpectral2011} induced by complex poses, clothing, etc. Another very recent work proposed in~\cite{SilNet2017} uses multi-view silhouette images to obtain reconstructions of free form blob-like objects/sculptures. However, the use of silhouettes limits the application to scenarios where background subtraction is assumed to be trivial. All these initial efforts are focused on textureless 3D reconstruction and do not seem to directly extend-able to non-rigid human body shapes. DynamicFusion~\cite{newcombe2015dynamicfusion} proposed the first dense SLAM system capable of reconstructing non-rigidly deforming scenes in real-time, by fusing together RGBD scans captured from commodity sensors. Similar to them, we shall be training our network with data from RGB/D sensors, however, unlike them, our pipeline shall be capable of high quality reconstruction from a single image by exploiting human body symmetry. Just as is true in their case, the reconstruction quality improves with providing more view information.

%Similarly, initial attempts in texture recovery have done so either for a small deformation space such as that of birds~\cite{texMalik} or require a large number of views~\cite{VideoSMPLTexture} and hence aren't directly extend able to our calibration-free single view 3D non-rigid reconstruction and texture recovery .
In regard to texture generation, recent work on multi-view synthesis \cite{DBLP:journals/corr/ZhaoWCLF17,AppearanceFlow} propose to generate an image from an alternate view-point given an input image. Recent work proposed in~\cite{texMalik} attempts texture recovery for category specific 3D models. However, the texture map is predicted only for the mean shape in the UV space and shown on reconstruction of birds. Nevertheless, this is not easily generalizable to human body shapes as accurate warping is non-trivial due to a large space of shape and pose variations associated with the human body.
%also criticise the idea of canonical space
Very recently, a model based learning paradigm outlined in~\cite{VideoSMPLTexture} proposed to generate human body models by estimating their deviation from the nearest SMPL model. Nevertheless, to recover the texture, the person is assumed to be seen from all the sides in a video sequence. Additionally such model based methods fail to deal with large geometric deformations induced by free-form clothing (e.g., rob or long skirt) scenarios.  
%(see Section~\ref{sec:related} for more details).    

\subsection{Our Contribution}
In this paper, we propose a deep learning based solution for textured 3D reconstruction of human body shapes given an input RGB image, in a calibration-free environment. 
%This is achieved by first recovering the volumetric shape of non-rigid human body shapes given a single view RGB image (at test time) followed by orthographic textured view synthesis using the respective depth projection of reconstructed (volumetric) shape and input RGB image. 
Given a single view RGB image, both reconstruction and texture generation are ill-posed problems. Thus, we proposed to co-learn the depth cues (using depth images obtained from affordable sensors like Kinect) with RGB images while training the network. This helps the network learn the space of complex body poses, which otherwise is difficult with just 2D content in RGB images. %This is achieved by intrinsic fusion of these two complementary modalities through common filter weights.
Although we propose to learn the reconstruction network with multi-view RGB and depth images (shown one at a time during training), co-learning them with shared filters enabled us to recover 3D volumetric shapes using just a single RGB image at test time. Apart from the challenge of non-rigid poses, the depth information also helps addressing the challenges caused by cluttered background, shape variations and free form clothing. 
Our texture recovery network uses a variational auto-encoder to generate orthographic texture images of reconstructed body models that are subsequently backprojected to recover a texture 3D mesh model. 
We show quantitative and qualitative results on four publicly available datasets (by simulating the RGB and D whenever unavailable) as well as real RGBD data collected with calibrated multi Kinect setup.  
%Motivate our solution in the context of challenges posed and existing work
The key contributions of this work are: 
\begin{itemize}
\item 
First, we introduce a novel deep learning pipeline to obtain textured 3D models of non-rigid human body shapes from a single image. To the best of our knowledge, obtaining the reconstruction of non-rigid shapes in a volumetric form (whose advantages we demonstrate) has not yet been attempted in literature. Further, this would be an initial effort in the direction of single view non-rigid reconstruction and texture recovery in an end-to-end manner.     %(Section~\ref{sec:method}) and show reconstruction results obtained using single view RGB input image. %To the best of our knowledge, both, non-rigid human body reconstruction as well as texture recovery of non-rigid 3D has not been attempted in a calibration-free environment using deep learning. 
\item Second, we demonstrate the importance of depth cues ({\em used only at train time}) for the task of non-rigid reconstruction. This is achieved by our novel training methodology of alternating RGB and D in order to capture the large space of pose and shape deformation. %This enables us to reconstruct 3D human body models from only signle view RGB image at test time. 
\item Third, we show that our model can partially handle non-rigid deformations induced by free form clothing, %by performing subject specific fine-tuning (Section~\ref{sec:expt}) 
as we do not impose any model constraint while training the volumetric reconstruction network.
\item Fourth, we proposed to use depth cues for texture recovery in the variational auto-encoder setup. This is the first attempt to do so in texture synthesis literature. 
\item Finally, we collected a real dataset (that shall be publicly released) of textured 3D human body models and their corresponding multi-view RGBD, that can be used in solving a variety of other problems such as human tracking, segmentation etc.     
%[Should we talk about proposing manually coloured dataset?]
\end{itemize}

\begin{figure*}[t!]
\begin{center}
\includegraphics[width=1.0\textwidth]{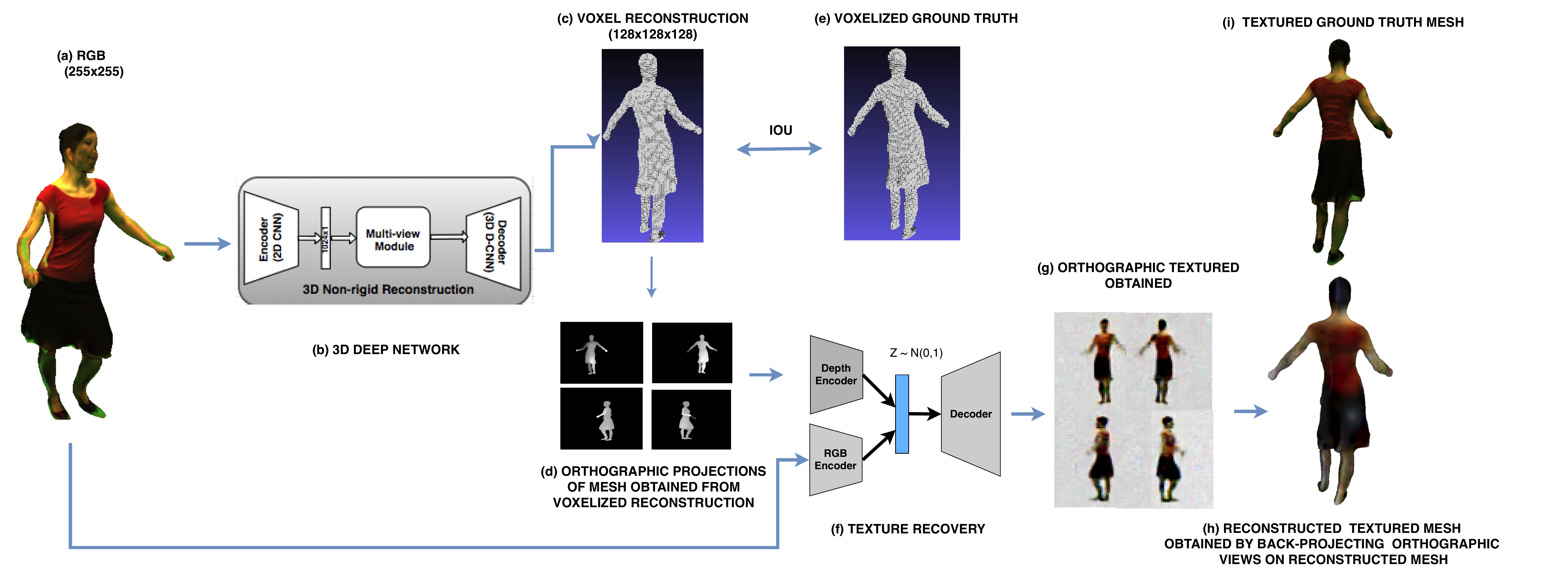}\\[1em]   
\caption{Proposed end-to-end deep learning pipeline for reconstructing textured non-rigid 3D human body models. Using a single view perspective RGB image (a), we perform a voxelized 3D reconstruction (c) using the reconstruction network (b). Then, to add texture to the generated 3D model, we first convert the voxels to a mesh representation using Poisson's surface reconstruction algorithm~\cite{kazhdan2013screened} and capture its four orthographic depth maps (d). These are fed as an input to the texture recovery network (f), along with the perspective RGB views used for reconstruction (a). The texture recovery network produces orthographic RGB images (g), that are back-projected onto the reconstructed model, to obtain the textured 3D model (h).}
\label{fig:pipeline}
\end{center}
\end{figure*}

%\subsection{Texture Recovery}
%Multi-view synthesis \cite{DBLP:journals/corr/ZhaoWCLF17,AppearanceFlow} is a unique application in computer vision where a novel view is synthesized given an image from another view. We propose a unique application of multi-view synthesis for recovering the texture of non-rigid 3D objects from a single image. \cite{texMalik} proposes texture recovery of category specific 3D models. However, the texture map is predicted only for the mean shape in the UV space, whose accurate warping is non-trivial for human body models which exhibit large shape and pose variations. Textured human body model is generated in \cite{VideoSMPLTexture} by estimating their deviation from nearest SMPL model. Nevertheless, to recover the texture, the person is assumed to be seen from all sides in a video sequence. 

\section{Our Method }
\label{sec:method}

A naive method of obtaining a textured 3D model would be to predict a colored occupancy grid. Given that the size of the volumetric output is $128^3$ and that for each such voxel, we would have to predict three values - one each for R, G and B (each in the range of $0-255$), it would be extremely computationally expensive to follow this method. Hence, we break down our end-to-end pipeline into two stages - one for 3D reconstruction and another for texture synthesis.     

Figure~\ref{fig:pipeline} shows the test time flow of the proposed end-to-end pipeline for reconstruction of textured non-rigid 3D shapes. A detailed outline of each stage is given below.

\subsection{Non-rigid Reconstruction}
%Several recent deep learning methods~\cite{3DR2N22016,SilNet2017,RayMalik2017} for class-specific rigid object reconstruction typically follow an encoder-decoder pipeline which consists of a convolutional and a deconvolution network respectively. 
For non-rigid reconstruction, we propose to use an encoder-decoder network with two methods for combining multi-view information - i) a 3D GRU (like in~\cite{3DR2N22016}) and ii) max pooling of the CNN feature vectors of the views (like in~\cite{SilNet2017}). The above two settings are considered to show that irrespective of the method of combining multi-view information for non-rigid human body reconstruction, depth information immensely help to capture the vast range of complex pose and shape variations while enhancing the reconstruction accuracy. Moreover, this provides the added advantage of obtaining 3D reconstruction from either only RGB or only D at test time, as discussed below. Further, by randomizing the number of views used at train time, we enable the network to reconstruct from a single image at test time. Intuitively, the network exploits the symmetry associated with non-rigid human shapes to reconstruct their 3D, even from a single view. \\

\noindent{\textbf{Network Architecture. }}  \\
a) \textit{Encoder - }ResNet-18 is used here that takes images of size 255x255 in one of the input modes (see below) and produces a 1024 dimensional feature vector. Each view produces one such feature vector, which is combined in the mulit-view module. We use Leaky ReLU as the activation function in both the encoder and decoder. \\
b) \textit{Multi-view Module - }Multi-view information is combined using either a max-pooling of the 1D CNN feature vectors, or using a 3D-GRU. The outputs are resized to $4^3$ and fed to the decoder\\
c) \textit{Decoder - } Deconvolutional ResNet-18 is used here that up-samples the output of the multi-view module to $128^3$\\

\noindent{{\textbf{Input Modes. }}In order to capture the large space of complex pose and shape deformations of humans, we experiment with four input modes: \\ 
a) \textit{RGB - }This setup is commonly used in rigid body reconstructions~\cite{RayMalik2017,3DR2N22016}. However, we qualitatively and quantitatively show in Section~\ref{sec:expt_recon} that this setup in inadequate for reconstructing non-rigid shapes.  \\
b) \textit{D - }The premise behind this mode is that depth-maps give us information about the geometry of the object, which as seen in Section~\ref{sec:expt_recon}, help in significantly enhancing the reconstruction quality. \\
c) \textit{RGBD - }In order to exploit both the depth and color channels, we augment RGB with D in a 4 channel input setup.\\
d) \textit{RGB/D - }Lastly, we propose a unique training methodology that gives us superior performance than the above 3 modes. Here, at train time, each mesh is reconstructed from either only multi-view RGB or multi-view depth. Thus, while we train with depth information, we can test with only RGB as well, which is a major advantage. Intuitively, this strategy is equivalent to sharing weights between the RGB and D spaces, in order to exploit the coherence between RGB and D; thus combining the advantages from both the spaces.  \\

\noindent{{\textbf{Loss Function. }}We use Voxel-wise Cross Entropy to train the reconstruction models. It is the sum of the cross-entropies for each pair of predicted and target voxel values. Let '$p$' be the predicted value at voxel $(i, j, k)$ and '$y$' the corresponding expected value. Then, the loss is defined as :

  \begin{equation}
  \begin{aligned}
  &L(p, y) = &\sum_{i, j, k} y_{(i, j, k)} \text{log}(p_{(i, j, k)}) + (1 - y_{(i, j, k)})\text{log}(1 - p_{(i, j, k)})
  \end{aligned}
  \end{equation}

\subsection{Texture Recovery}

We obtain a set of four orthographic depth maps (excluding top and bottom views) of the 3D human mesh $M$ generated by the reconstruction network and represent them as $D = \{\pi_{D,i}^{o}\}$. Here $\pi_{D,i}^{o}$ corresponds to the orthographic ($o$) depth projection ($D$) of $M$ to the $i^{th}$ face of the cube. From the set of perspective RGB images used for reconstruction $P = \pi_{RGB,j}^{P}$, we randomly choose an image as an input to the texture recovery network. Given each of the orthographic depth images and a randomly chosen perspective image, texture recovery aims to estimate the color of each pixel in the depth map. \\

In order to achieve this, we propose a simple texture recovery approach that detaches the requirement of a calibrated setup by employing a variational auto-encoder (VAE). Another generative model, generative adversarial networks (GAN) can be employed, however, we leverage upon VAEs since it is better at understanding the global shape of the object while GANs are traditionally used for recovering finer details. The VAE is trained by maximizing the log likelihood $log(p_\theta(\pi_{RGB,i}^{O} | \pi_{D,i}^{O}, \pi_{RGB,j}^{P})$ which is equivalent to minimizing the variational lower bound, as indicated by the equation below. The VAE is trained to learn the distribution $p(\pi_{RGB,i}^{O} | \pi_{D,i}^{O}, \pi_{RGB,j}^{P})$ which models the color of the provided orthographic depth maps. 
\begin{equation}
\begin{aligned}
V(\pi_{RGB,i}^O, \pi_{D,i}^O,\pi_{RGB,j}^P;\theta,\phi) &= -KL(q_{\phi}(z|\pi_{D,i}^{O}, \pi_{RGB,j}^{P}) || p_\theta(z) )\\
&+ E_{q_\phi}[logp_\theta(\pi_{RGB,i}^{O})|\pi_{D,i}^O,\pi_{RGB,j}^P,z] 
\end{aligned}
\end{equation}
The data is modeled by normalizing out $z$ from the joint probability distribution $p(\pi_{RGB,i}^{O},z)$. Further, $p(z)$ is inferred as $p_\theta(z |\pi_{D,i}^{O}, \pi_{RGB,j}^{P})$ and is modeled by the encoder, parameterized by $\theta$. Variational distribution $q_{\phi}$ is introduced to approximate the unknown true posterior $p_\theta$. For further details, refer to the formulation of Variational Auto-encoders ~\cite{kingma2013auto}.The second term in equation 2 is the log likelihood of samples, which is the L2 reconstruction loss that encourages consistency between the encoder and decoder. The KL divergence reduces the distance between the variational distribution and prior distribution. $q_\phi(z|x)$ is associated with a prior distribution over the latent variables for which we use a multivariate Gaussian with unit variance $\mathcal{N} (0,I)$. To the best of our knowledge, this is the first attempt to use depth cues for novel view synthesis. \\

\noindent{{\textbf{Network Architecture. }}As shown in Figure~\ref{fig:pipeline}, the encoder consists of two symmetric branches, one each, for orthographic depth, and perspective RGB image as input. However, the weights are not shared between the two encoders. These networks consist of convolutional layers which have 64,128,256,256,512 and 1024 channels with filters of size $5$,$5$,$3$,$3$,$3$ and $4$ respectively followed by a fully connected layer with 1024 neurons. The representations from the two branches are combined to form a 1024 dimensional latent variable. The decoder network consists of a fully connected layer with $256\times8\times8$ neurons. Deconvolution layers are then followed with $2\times2$ upsampling and have 256, 256, 256, 128, 64 and 3 channels. The filter sizes are $5\times5$ for all layers. ReLU is used as the activation function and the output size is set to $64 \times 64$.

\section{Experiments \& Results}

%In this section, we corroborate our method of single view non-rigid 3D reconstruction and texture recovery by means of qualitative and quantitative evaluation. On the reconstruction forefront, we show the importance of depth cues for the non-rigid setup as well as the improvement in performance with our unique learning paradigm of sharing the weights for both the RGB and D spaces. Further, we show that our model is capable of learning non-rigid deformations induced by clothing, without explicitly learning a separate model for clothing. On the texture recovery forefront, we show that the orthographic projections can be colored with single RGB perspective image. This unique approach enables to recover texture from single view in a calibration free set up.

\subsection{Datasets}
\label{sec:datasets}
\noindent{\bf MPI Datasets:}  We use two datasets from MPI. First, we use the parametric SMPL model~\cite{SMPL2016,surreal} to generate 10 mesh sequences, each containing 300 frames, i.e., 3000 meshes, consisting of an equal number of male and female models. Additionally, we use FAUST's data ~\cite{Bogo:CVPR:2014} which consists of 300 high resolution human meshes.%, each having approximately 250,000 vertices. Each scan is a high-resolution, triangulated, non-watertight mesh acquired with a 3D multi-stereo system. 
 There are a total of 10 different subjects in 30 different poses. The 300 meshes come divided in 2 sequences, one having complete meshes and the other having broken/incomplete parts - the former used for training, and the latter for testing. 
In addition, we simulated a virtual Kinect setup to capture aligned RGBD. Since both these datasets had correspondences, a few meshes were manually textured, and this texture was transferred across each of the datasets. \\

\noindent{\bf MIT's Articulated Mesh Animation \cite{VlasicMIT2008}: } This dataset consists of 5 mesh sequences (approx. 175 to 250 frames each). It provides RGB images from 8 views and the corresponding 3D meshes for each frame. The total number of meshes used from this dataset for training are 1,525. \\ 
%In order to test our texture recovery, we manually backprojected colour onto meshes from the 'Samba' sequence.
\newline
\noindent{\bf Our Data:} A total of 5 mesh sequences, each containing 200 to 300 frames with significant pose and shape variation were captured using a calibrated multi-Kinect setup. The colored point clouds were re-meshed using Poisson's surface reconstruction algorithm, after pre-processing them for noise removal. The processed RGBD and corresponding textured models were used for training the pipeline. %The dataset and associated code shall be publicly released.  \\
\begin{figure*}[bt!]
\centering
\subfigure{
\begin{adjustbox}{width=0.88\textwidth}
\centering
\begin{tabular}{M{0.8cm}M{2cm}M{2cm}M{2cm}M{2cm}M{2cm}M{2cm}M{2cm}M{1.9cm}}
OURS & \subfigure{\includegraphics[width=2cm, height=2.5cm]{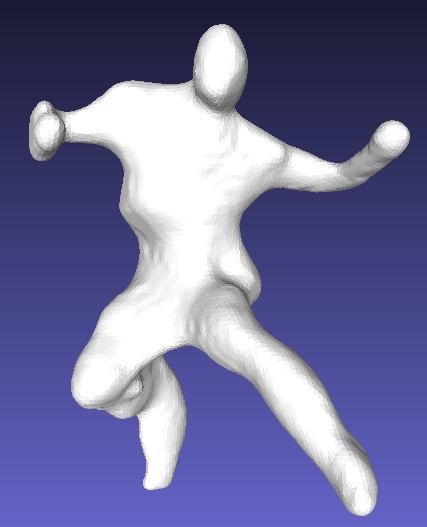}} &
\subfigure{\includegraphics[width=2cm, height=2.5cm]{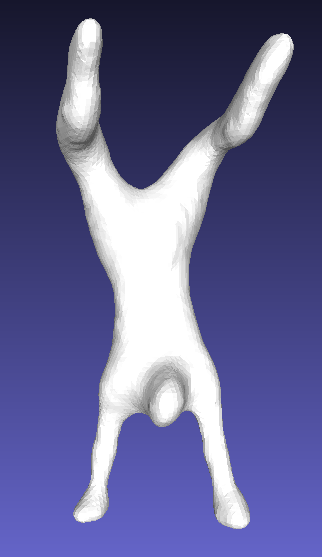}} & 
\subfigure{\includegraphics[width=2cm, height=2.5cm]{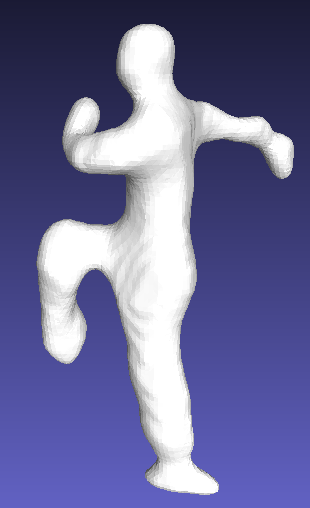}} &
\subfigure{\includegraphics[width=2cm, height=2.5cm]{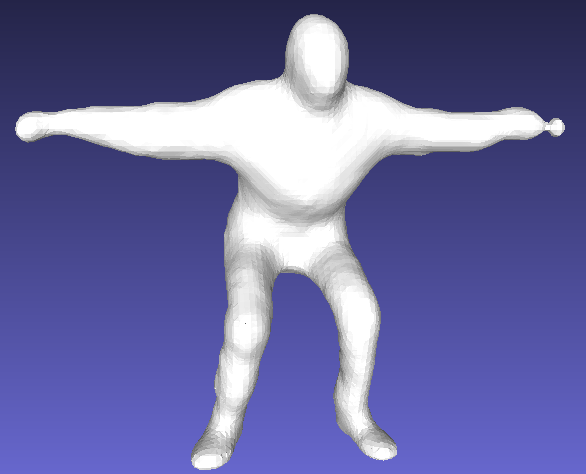}} & \subfigure{\includegraphics[width=2cm, height=2.5cm]{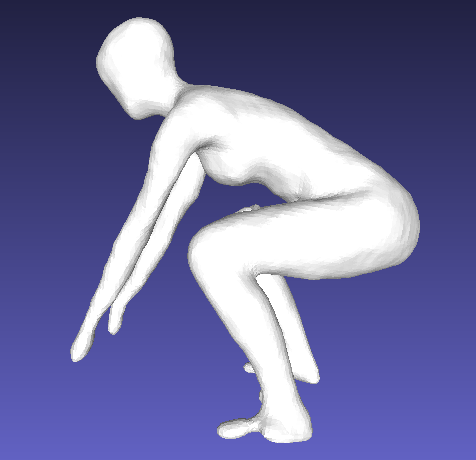}} &
\subfigure{\includegraphics[width=2cm, height=2.5cm]{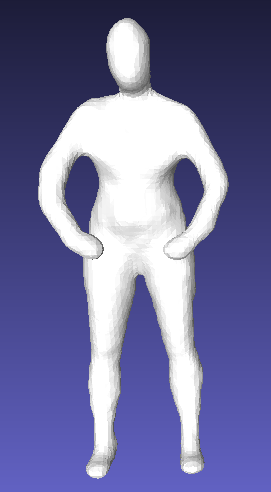}} & 
\subfigure{\includegraphics[width=2cm, height=2.5cm]{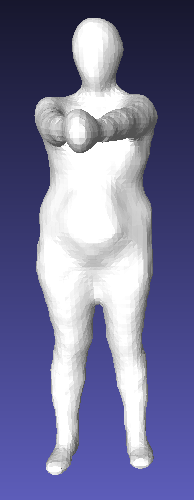}} &
\subfigure{\includegraphics[width=2cm, height=2.5cm]{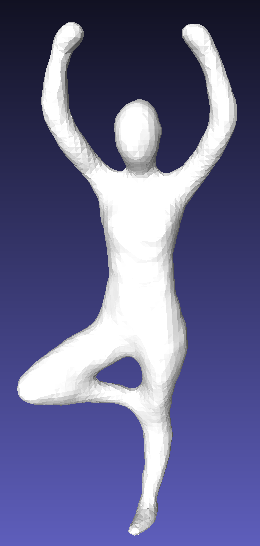}} \\
\quad
GT (MVG) & \subfigure{\includegraphics[width=2cm, height=2.5cm]{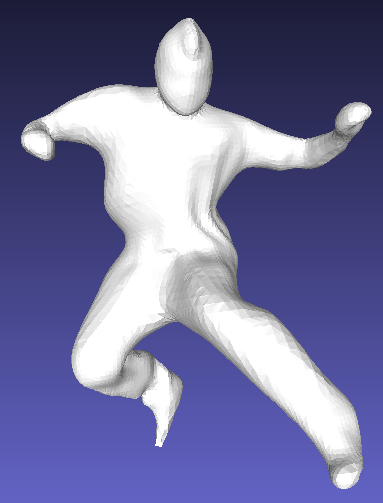}} &
\subfigure{\includegraphics[width=2cm, height=2.5cm]{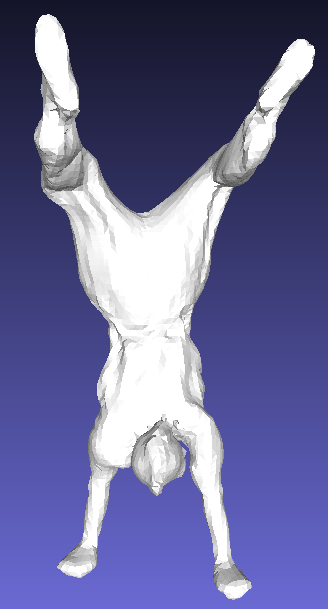}} & 
\subfigure{\includegraphics[width=2cm, height=2.5cm]{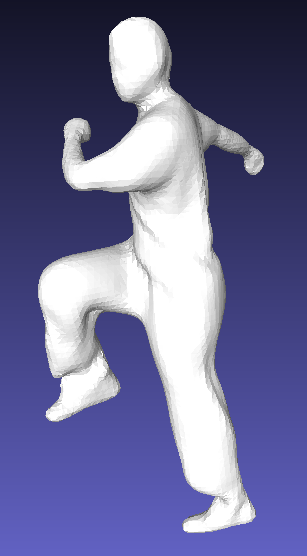}} &
\subfigure{\includegraphics[width=2cm, height=2.5cm]{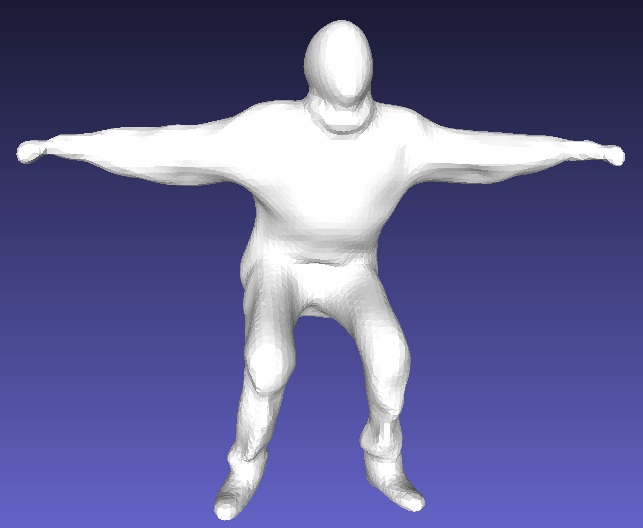}} & \subfigure{\includegraphics[width=2cm, height=2.5cm]{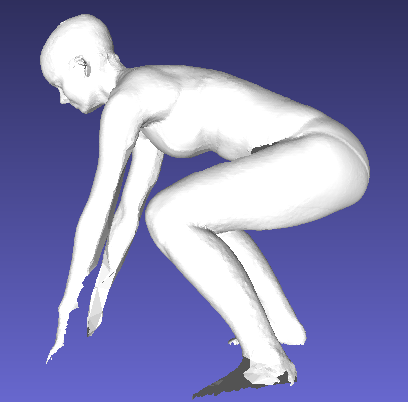}} &
\subfigure{\includegraphics[width=2cm, height=2.5cm]{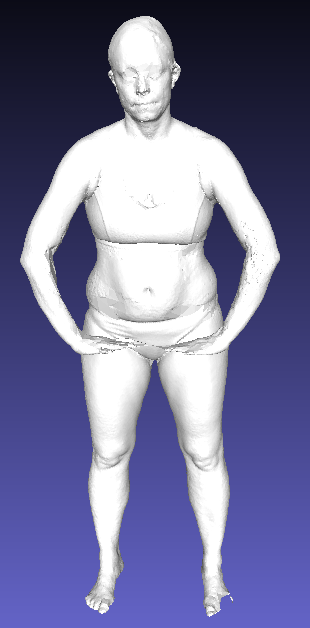}} & 
\subfigure{\includegraphics[width=2cm, height=2.5cm]{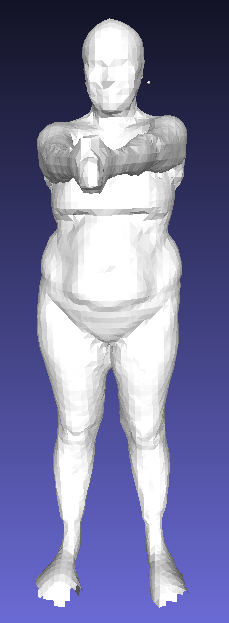}} &
\subfigure{\includegraphics[width=2cm, height=2.5cm]{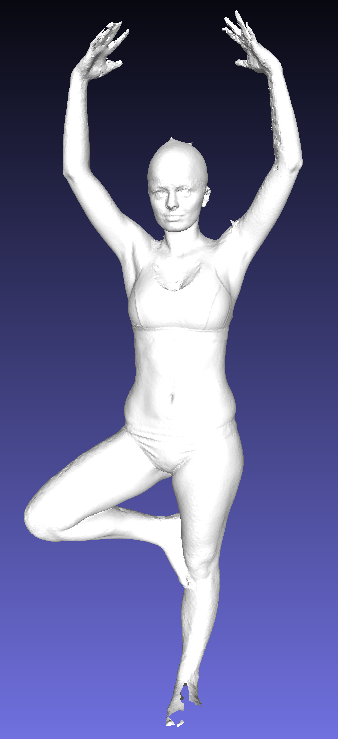}} \\
\end{tabular}
\end{adjustbox}
}
\caption{3D Shapes obtained with our reconstruction network compared to ground truth models obtained with MVG. }
\label{fig:recon}
\end{figure*}

\label{sec:expt}
\subsection{Non-rigid Reconstruction}
\label{sec:expt_recon}
\begin{table*}
\centering
\begin{tabular}{|c|c|c|c|c|c|}
\hline
Dataset & Multi-View & RGB (Baseline) & D & RGBD & RGB/D (Ours)\\
  \hline
\hline
\multirow{2}{*}{MPI-SMPL~\cite{surreal}} & 3D-GRU & 0.6903 & 0.7709 & 0.7541 & \bf{0.8040}\\
& Max Pool & 0.7144 & 0.7633 & 0.7550 & \bf{0.7816}\\ \hline
\multirow{2}{*}{MIT~\cite{VlasicMIT2008}} & 3D-GRU & 0.0103 & 0.7403 & - & \bf{0.7547}\\
& Max Pool & 0.0081 & 0.7205 & - & \bf{0.7480}\\ \hline
\multirow{2}{*}{MPI-FAUST~\cite{dfaust:CVPR:2017}} & 3D-GRU & 0.8113 & 0.8629 & 0.8356 & \bf{0.8644}\\
& Max Pool & 0.8150 & \bf{0.8661} & 0.8366 & 0.8521\\ \hline
\multirow{2}{*}{OUR DATA} & 3D-GRU & 0.6816 &  0.7963 & 0.8114 & \bf{0.8241}\\
& Max Pool & 0.6883 & 0.7844 & 0.8017 & \bf{0.8066}\\ \hline
\end{tabular}
\caption{A comparison of IoU values tested using a single view on datasets~\cite{VlasicMIT2008,surreal,dfaust:CVPR:2017}, under the various input modes, when trained with two different view modules.}
\label{table:iou}
\end{table*}

\noindent{\textbf{Network's Training. }}  We used Nvidia's GTX 1080Ti, with 11GB of VRAM to train our models. A batch size of 5 with the ADAM optimizer having an initial learning rate of $10^{-4}$, and a weight decay of $10^{-5}$ is used to get optimal performance. Further, a standard $80:20$ split between the training and testing datasets is adhered to. In order to ensure that reconstruction is feasible from single as well as multiple views, we choose random number of views from available views for training a mesh in each iteration. Using this randomization in training, we are providing sufficient view information to the network so that it can learn the space of body pose and shape variations and hence able to achieve single view reconstruction at test time.  \\

\noindent{\textbf{Evaluation Metric. }} The primary metric used to evaluate our performance is the Intersection over Union (IoU), which is a comparison between the area of overlap and the total area  encompassing both the objects. Larger its value, the better the quality of reconstruction. 

Let '$p$' be the predicted value at voxel $(i, j, k)$ and '$y$' the corresponding expected value. '$I$' is an indicator function which gives gives a value of 1 if the expression it is evaluating is true, if not, it gives 0. '$t$' is an empirically decided threshold of 0.5 above which the cell is considered as filled.

\begin{equation}
  IoU = \frac{\sum_{i, j, k} [I(p(i, j, k) > t) I(y(i, j, k))]}{\sum_{i, j, k} [I(p(i, j, k) > t) + I(y(i, j, k))]}
  \end{equation}
%  \begin{wrapfigure}{l}{0.3\textwidth}
% \includegraphics[width=4cm, height=5cm]{./abbhinav-combine-figures1.pdf}
% \caption{Clothing induced deformations captured by our proposed method on~\cite{VlasicMIT2008}.}
% \centering
% \label{fig:cloth}
% \end{wrapfigure}
\begin{figure*}[h!]
\begin{center}
\includegraphics[width=2cm, height=2.8cm]{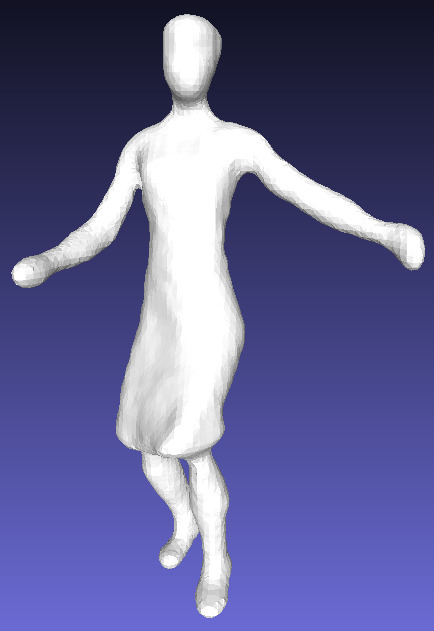} \includegraphics[width=2cm, height=2.8cm]{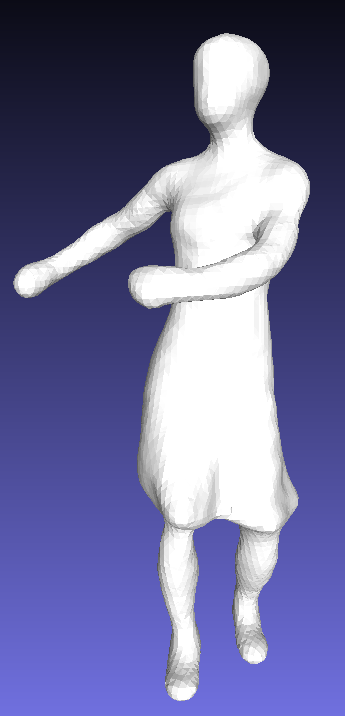} \includegraphics[width=2cm, height=2.8cm]{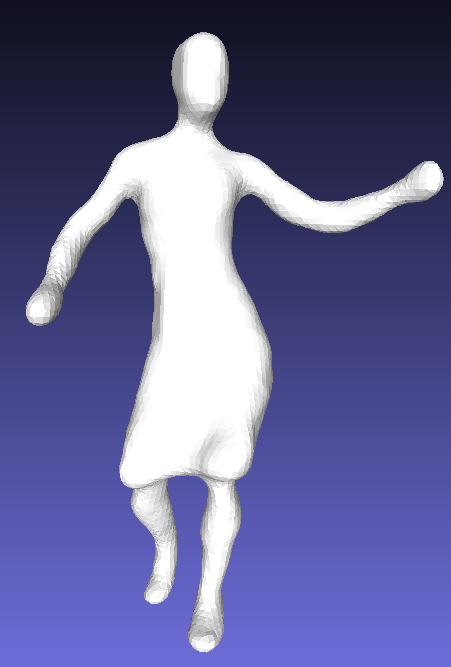} \includegraphics[width=2cm, height=2.8cm]{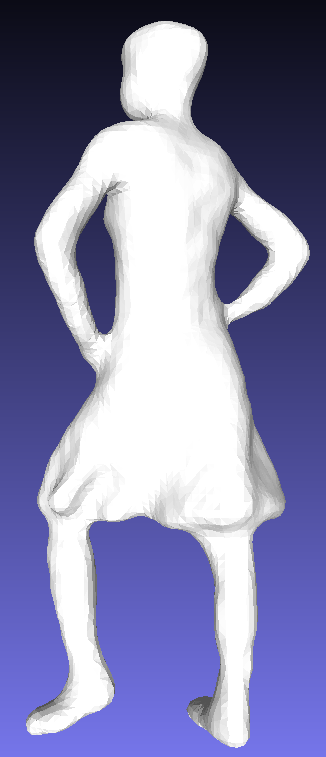} \caption{Clothing induced deformations captured by our proposed method on~\cite{VlasicMIT2008}.}
\label{fig:cloth}
\end{center}
\end{figure*}

\noindent{\textbf{Baseline. }}As described in Section~\ref{sec:related}, there are several standard encoder-decoder networks that use single/multi-view RGB image(s)  for voxelized reconstruction of rigid objects. Therefore, as baseline, we interpret this setting of using only RGB image(s) for both training and testing the reconstruction network. Further, the qualitative and quantitative results are compared and evaluated on ground truth data generated using  traditional Multi-View Geometry (MVG) setups (from upto 22 views), on four datasets of varying complexity. For rendered sample outputs, please refer to the supplementary video. \\

%\noindent{\textbf{Analysis. }}
{\noindent \textbf{Results \& Discussion. }}Quantitative results (IOU metric) in Table~\ref{table:iou} suggest that for a variety of datasets of varying complexity and irrespective of the method of combining multiple views, the depth information is very critical for accurate reconstruction of human models. It is interesting to notice that the difference in IoU values between RGB and RGB/D widens under two scenarios - a) when the dataset has very complicated poses (such as the handstand sequence in MIT) and b) when the background becomes more complicated. Figure~\ref{fig:back} re-emphasizes this ineffectiveness of using RGB alone, in which the first and second rows show the reconstructions from MIT's handstand and our captured data, respectively.  An intuition behind the working of this training paradigm is that the co-learning of the shared filter weights of the two modalities act as a regularization for one another, thus enhancing the information seen by the network. \\
% \begin{wrapfigure}{r}{6cm}
% \subfigure{
% \begin{adjustbox}{width=0.44\textwidth}
% \centering
% \begin{tabular}{M{1.8cm}M{1.8cm}M{1.8cm}M{1.8cm}}
% Input RGB & Ground Truth & RGB Reconstruction & Ours (RGB/D) \\
% \subfigure{\includegraphics[width=2cm, height=2.5cm]{./rgbd_cmp/hs_103_ip}} & 
% \subfigure{\includegraphics[width=2cm, height=2.5cm]{./rgbd_cmp/hs_103_gt}} &
% \subfigure{\includegraphics[width=2cm, height=2.5cm]{./rgbd_cmp/hs_103_d}} & 
% \subfigure{\includegraphics[width=2cm, height=2.5cm]{./rgbd_cmp/hs_103_rgbd}} \\
% \quad
% \subfigure{\includegraphics[width=2cm, height=2.4cm]{./rgbd_cmp/sagar_pers}} & 
% \subfigure{\includegraphics[width=2cm, height=2.5cm]{./rgbd_cmp/sagar_gt}} &
% \subfigure{\includegraphics[width=2cm, height=2.5cm]{./rgbd_cmp/sagar_rgb}} & 
% \subfigure{\includegraphics[width=2cm, height=2.5cm]{./rgbd_cmp/sagar_rgbd}} \\
% \end{tabular}
% \end{adjustbox}
% }
% \caption{Qualitative comparison with baseline method.}
% \label{fig:back}
% \end{wrapfigure}
\begin{figure}
\centering
\begin{tabular}{M{1.8cm}M{1.8cm}M{1.8cm}M{1.8cm}}
Input RGB & Ground Truth (MVG) & RGB Reconstruction & Ours (RGB/D) \\
\subfigure{\includegraphics[width=2cm, height=2.5cm]{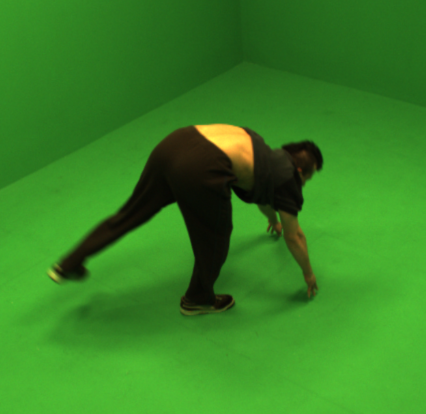}} & 
\subfigure{\includegraphics[width=2cm, height=2.5cm]{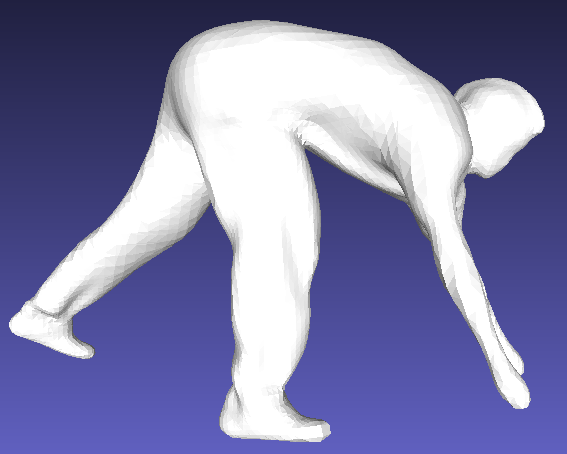}} &
\subfigure{\includegraphics[width=2cm, height=2.5cm]{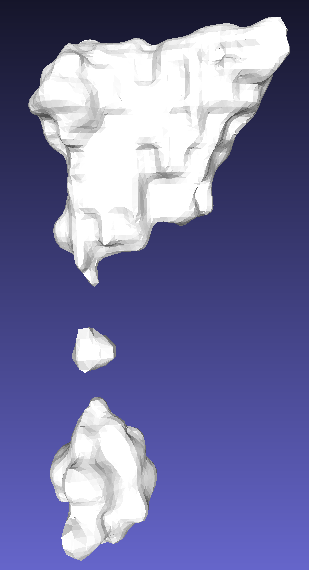}} & 
\subfigure{\includegraphics[width=2cm, height=2.5cm]{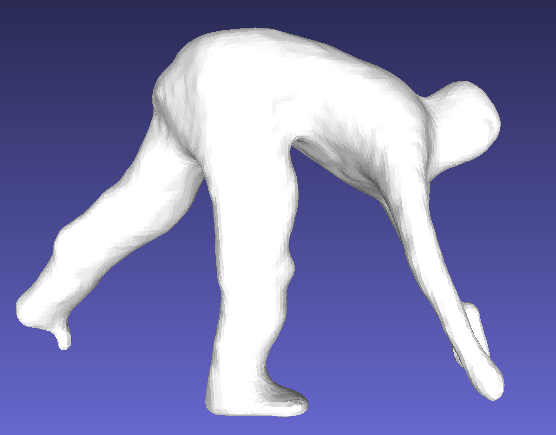}} \\
\quad
\subfigure{\includegraphics[width=2cm, height=2.4cm]{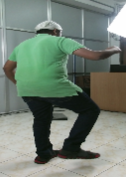}} & 
\subfigure{\includegraphics[width=2cm, height=2.5cm]{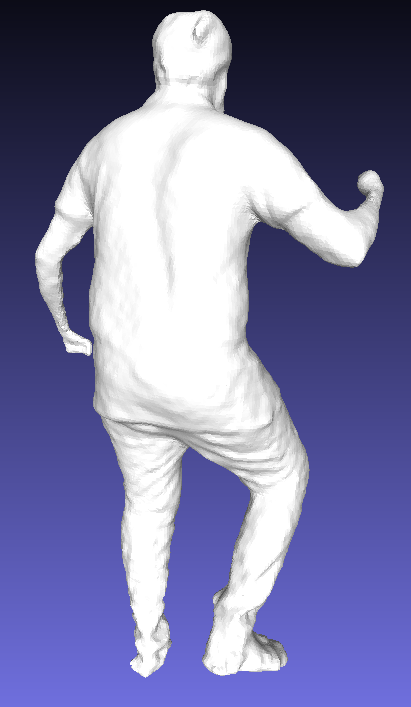}} &
\subfigure{\includegraphics[width=2cm, height=2.5cm]{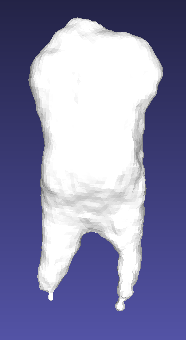}} & 
\subfigure{\includegraphics[width=2cm, height=2.5cm]{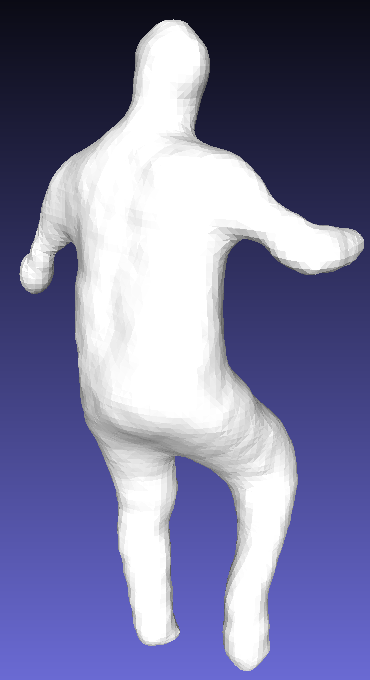}} \\
\end{tabular}
\caption{Qualitative comparison with baseline method.}
\label{fig:back}
\end{figure}
Figure~\ref{fig:recon} shows the robustness of the learned model performing a vast range of actions. This robustness while reconstructing from a single image can be attributed to the network's ability to exploit the symmetry associated with non-rigid human shapes.  %While model based approaches might successfully predict the closest human pose and shape, a voxel-grid based model enables obtaining low level geometrical details. This advantage of our approach is depicted in Figure~\ref{fig:cloth}. 
As a result of not imposing any body model constraint, we were able to partially handle non-rigid deformations induced by free form clothing as shown in Figure~\ref{fig:cloth}. While the current pipeline has not been trained to explicitly capture the temporal information available in sequences, results on MIT's Samba dance sequence ~\cite{VlasicMIT2008} shown in Figure~\ref{fig:cloth} show great promise for reconstructing performance capture scenarios from a single image. 
% \begin{wrapfigure}{r}{7cm}
% \subfigure{
% \begin{adjustbox}{width=0.54\textwidth}
% \centering
% \begin{tabular}{M{1.8cm}M{1.8cm}M{1.8cm}|M{1.8cm}M{1.8cm}M{1.8cm}}
% Input RGB & Ours (RGB/D) & GT & Input RGB & Ours (RGB/D) & GT\\
% \hline
% \subfigure{\includegraphics[width=2cm, height=2.5cm]{./recon_figs/march2_ip}} & 
% \subfigure{\includegraphics[width=2cm, height=2.5cm]{./recon_figs/march2_r}} & 
% \subfigure{\includegraphics[width=2cm, height=2.5cm]{./recon_figs/march2_gt}} &
% \subfigure{\includegraphics[width=2cm, height=2.5cm]{./recon_figs/march2_ip}} & 
% \subfigure{\includegraphics[width=2cm, height=2.5cm]{./recon_figs/march2_r}} & 
% \subfigure{\includegraphics[width=2cm, height=2.5cm]{./recon_figs/march2_gt}}
% \end{tabular}
% \end{adjustbox}
% }
% \caption{A few sample reconstructions on our real data, generated from single view images. }
% \label{fig:realrecon}
% \end{wrapfigure}
\subsection{Texture Recovery}
%\noindent{\textbf{Analysis. }}
The textured model is obtained by synthesizing a texture orthographic view image corresponding to respective orthographic depth image taken from volumetric model predicted by reconstruction module. \\ 

{\noindent \textbf{Results \& Discussion.}} Figure~\ref{fig:texture} shows the generated orthographic texture image and corresponding textured model obtained after their back-projection on reconstructed 3D model on samples from various datasets. For rendered textured meshes, please refer to the supplementary video. The variational autoencoder predicts low resolution images of size $64x64$. Attempting to predict higher resolution texture images using only this framework adds additional complexity to the network, and results in less accurate orthographic texture images. Further post-processing using super-resolution networks, usage of multi-view input images etc. can be used to improve the resolution of the synthesized texture images to perform high quality texture recovery. \\ 

Since we use just a single image to recovery the texture, the nature of the textures recovered shall be simple. Recovery of complex textures from only a single image is not a realistic expectation for any texture synthesis methodology. However, the task of single view texture recovery has practical applications in performance capture scenarios, AR/VR platforms etc., and this is an initial effort to address that gap. Given a prior on the type of clothing or sequence that is going to be recovered (for example, if we would like to recover the texture of a particular dance form), the proposed texture recovery pipeline shall intuitively perform better (as it currently trained for multiple different scenarios). 

%The generative model proposed in our architecture maps input depth images (obtained from orthographic project of reconstructed shape) to the closest image the network has seen during the train phase. 
%The proposed variational auto-encoder seems robust to noise in real capture set up as it generates the missing body parts. Since the approach is not model based, it can model to generate complex clothing deformations as shown in Figure~\ref{fig:texture}. The proposed method allows to incorporate multiple perspective images to recover complex textures.
\begin{figure*}[ht!]
\centering
\subfigure{
\begin{adjustbox}{width=0.75\textwidth}
\centering
\begin{tabular}{M{1.8cm}M{1.8cm}M{1.8cm}M{1.8cm} M{1.8cm}M{1.8cm}M{1.8cm}M{1.8cm}}
Perspective Image & Reconstruction & Reconstruction GT & view 1 & view 2 & view 3 & view 4 & Textured Model \\
\subfigure{\includegraphics[width=2cm, height=2.5cm]{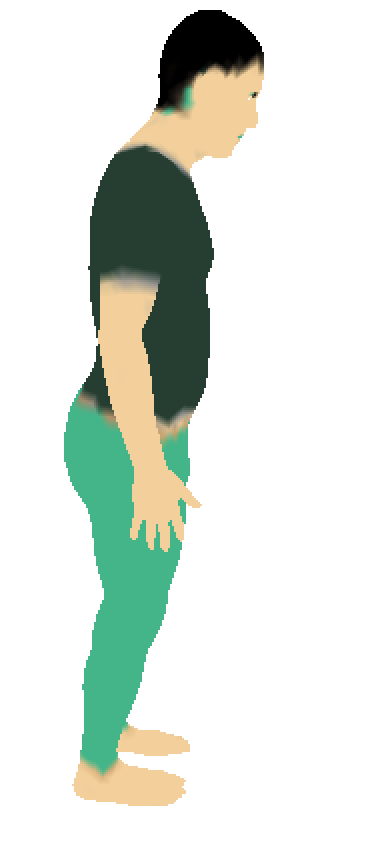}} &
\subfigure{\includegraphics[width=2cm, height=2.5cm]{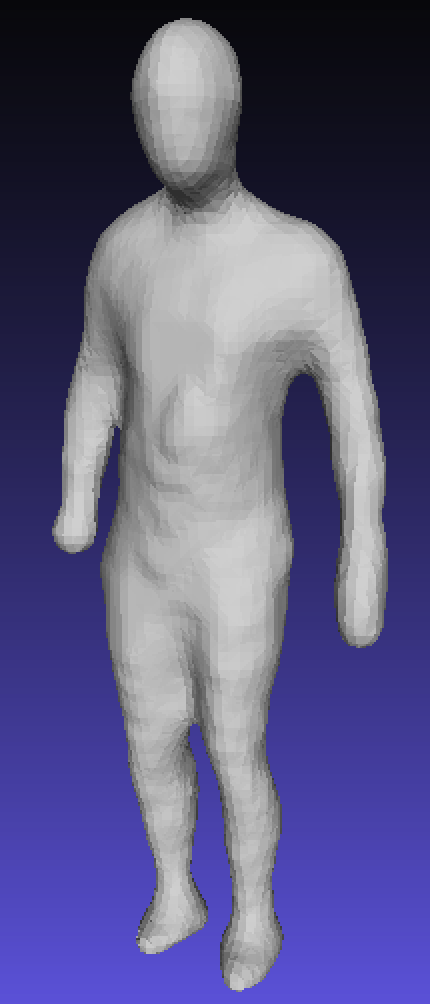}} & 
\subfigure{\includegraphics[width=2cm, height=2.5cm]{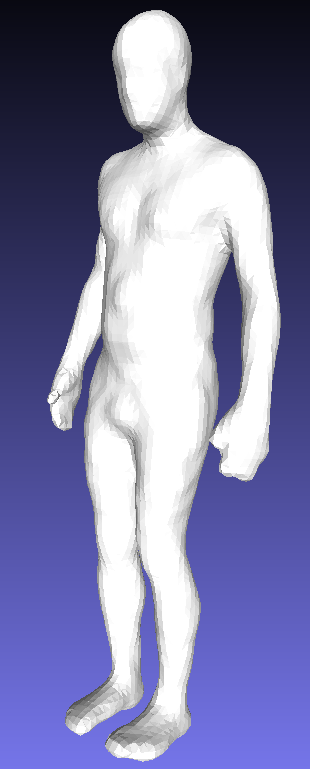}} &
\subfigure{\includegraphics[width=2cm, height=2.5cm]{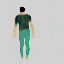}} &
\subfigure{\includegraphics[width=2cm, height=2.5cm]{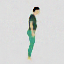}} & 
\subfigure{\includegraphics[width=2cm, height=2.5cm]{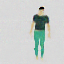}} &
\subfigure{\includegraphics[width=2cm, height=2.5cm]{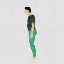}}&
\subfigure{\includegraphics[width=2cm, height=2.5cm]{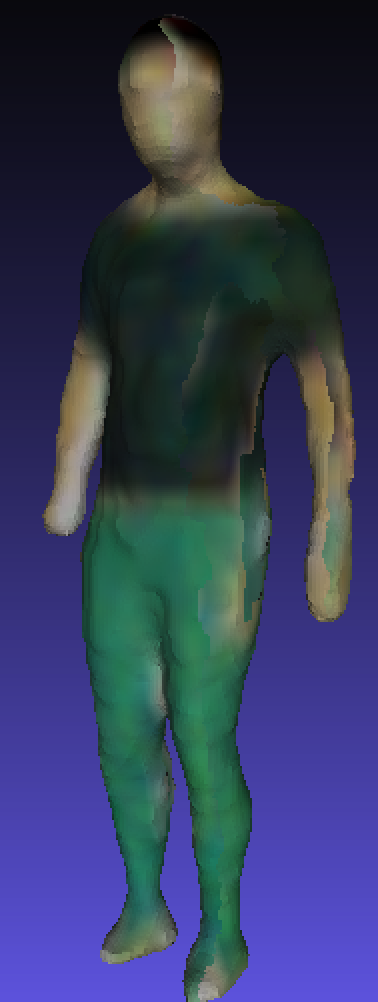}}\\
\quad

\subfigure{\includegraphics[width=2cm, height=2.5cm]{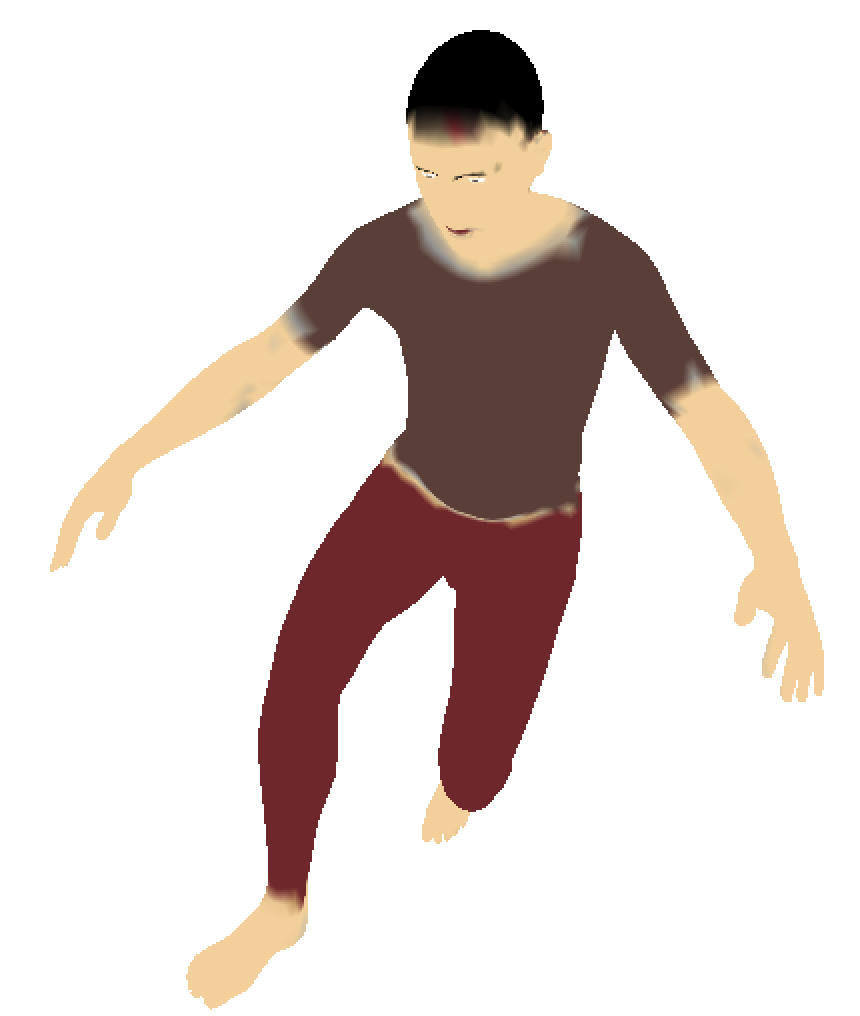}} &
\subfigure{\includegraphics[width=2cm, height=2.5cm]{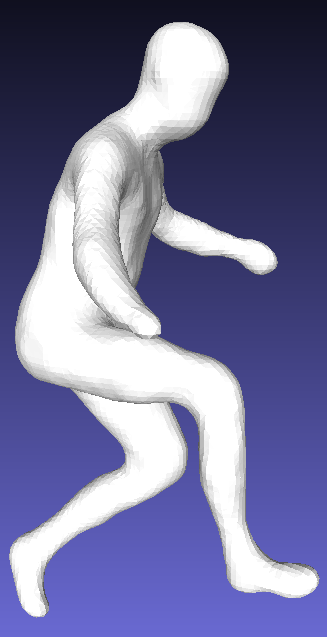}} &
\subfigure{\includegraphics[width=2cm, height=2.5cm]{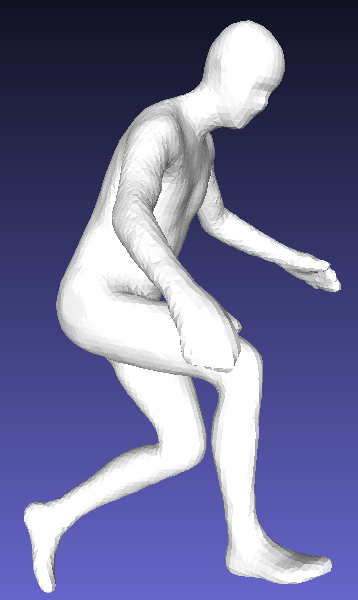}} &
\subfigure{\includegraphics[width=2cm, height=2.5cm]{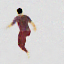}} &
\subfigure{\includegraphics[width=2cm, height=2.5cm]{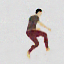}} &
\subfigure{\includegraphics[width=2cm, height=2.5cm]{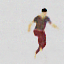}} &
\subfigure{\includegraphics[width=2cm, height=2.5cm]{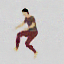}} &
\subfigure{\includegraphics[width=2cm, height=2.5cm]{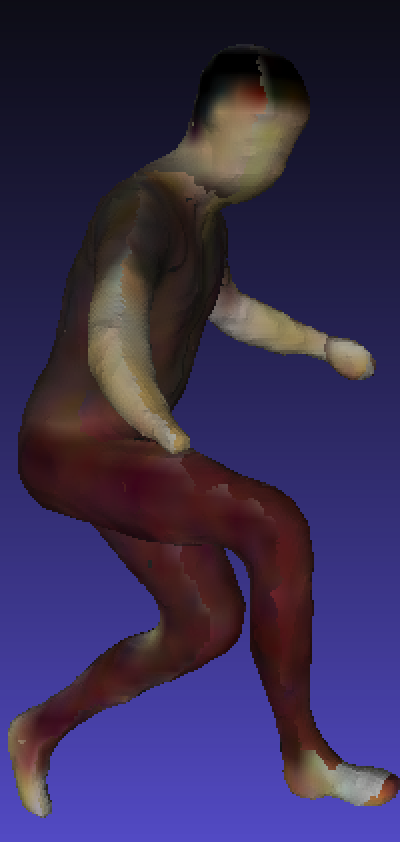}} \\
\quad
\subfigure{\includegraphics[width=2cm, height=2.5cm]{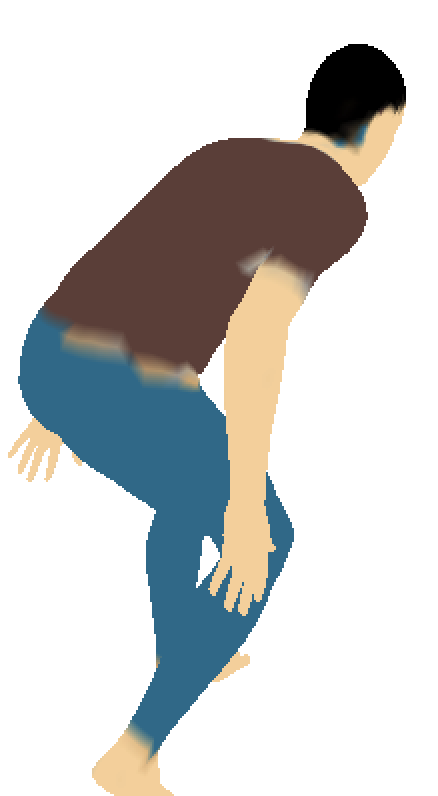}} &
\subfigure{\includegraphics[width=2cm, height=2.5cm]{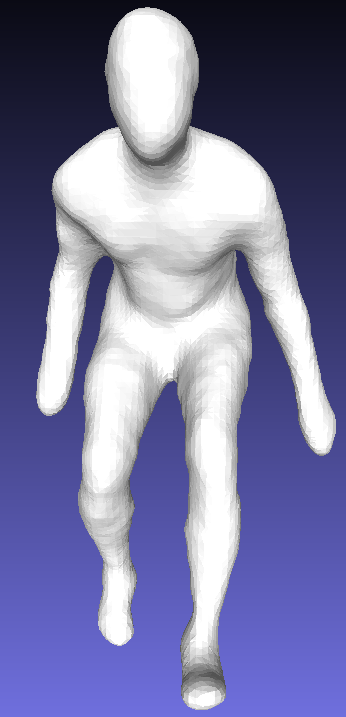}} &
\subfigure{\includegraphics[width=2cm, height=2.5cm]{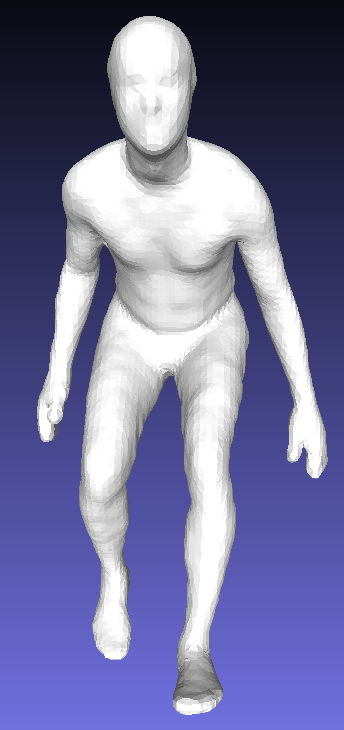}} &
\subfigure{\includegraphics[width=2cm, height=2.5cm]{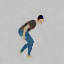}} &
\subfigure{\includegraphics[width=2cm, height=2.5cm]{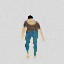}} &
\subfigure{\includegraphics[width=2cm, height=2.5cm]{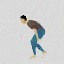}} &
\subfigure{\includegraphics[width=2cm, height=2.5cm]{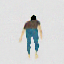}} &
\subfigure{\includegraphics[width=2cm, height=2.5cm]{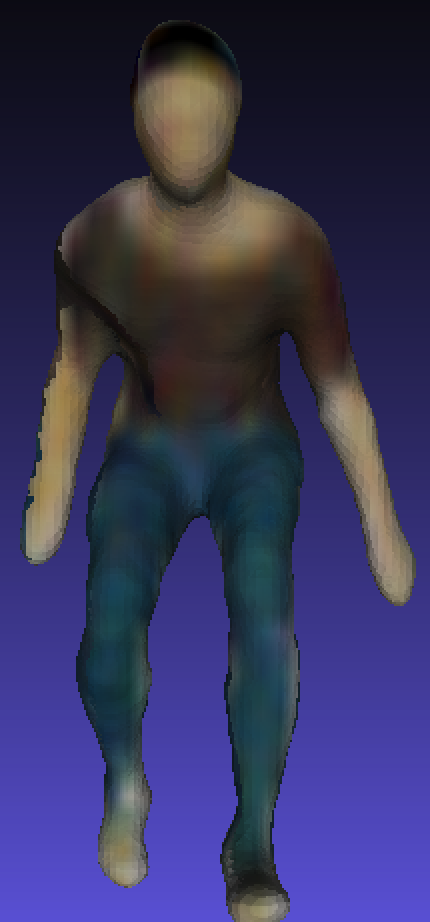}} \\
\subfigure{\includegraphics[width=2cm, height=2.5cm]{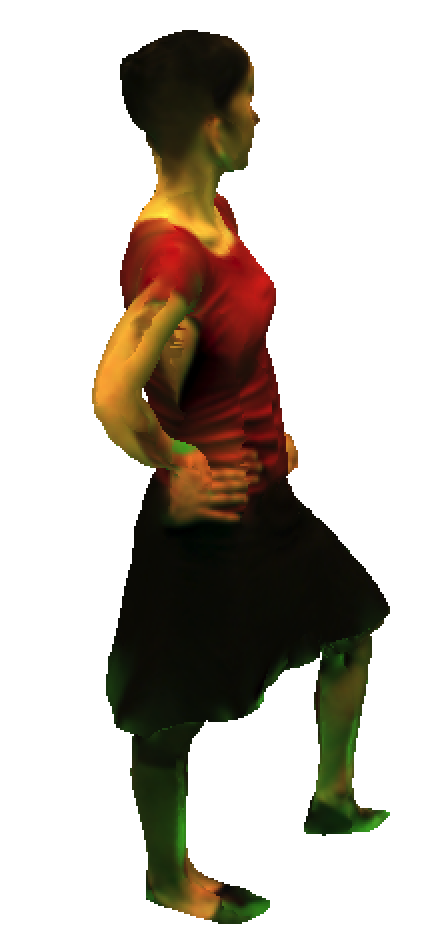}} &
\subfigure{\includegraphics[width=2cm, height=2.5cm]{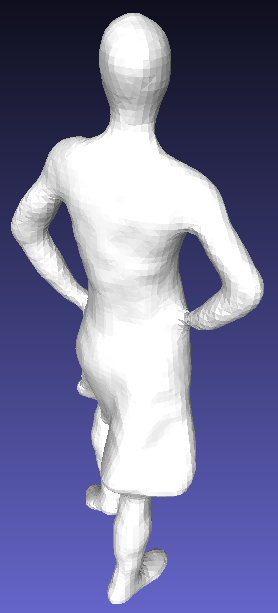}} &
\subfigure{\includegraphics[width=2cm, height=2.5cm]{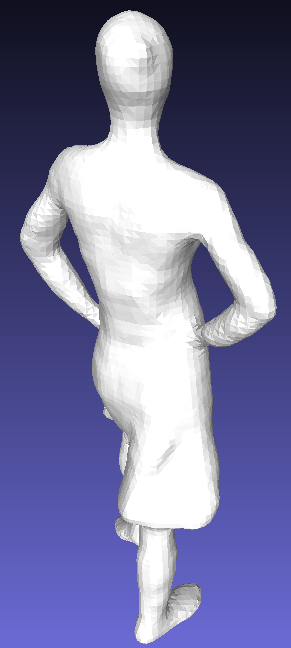}} &
\subfigure{\includegraphics[width=2cm, height=2.5cm]{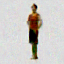}} &
\subfigure{\includegraphics[width=2cm, height=2.5cm]{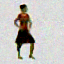}} &
\subfigure{\includegraphics[width=2cm, height=2.5cm]{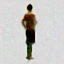}} &
\subfigure{\includegraphics[width=2cm, height=2.5cm]{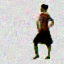}} &
\subfigure{\includegraphics[width=2cm, height=2.5cm]{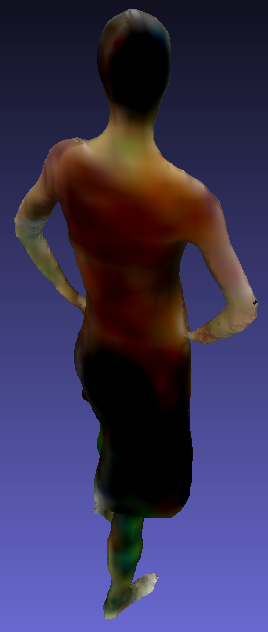}} \\
\subfigure{\includegraphics[width=2cm, height=2.5cm]{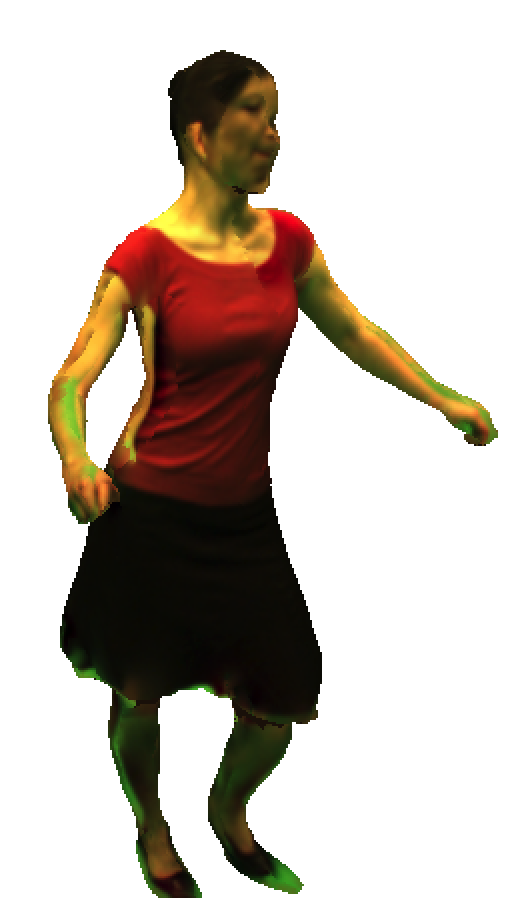}} &
\subfigure{\includegraphics[width=2cm, height=2.5cm]{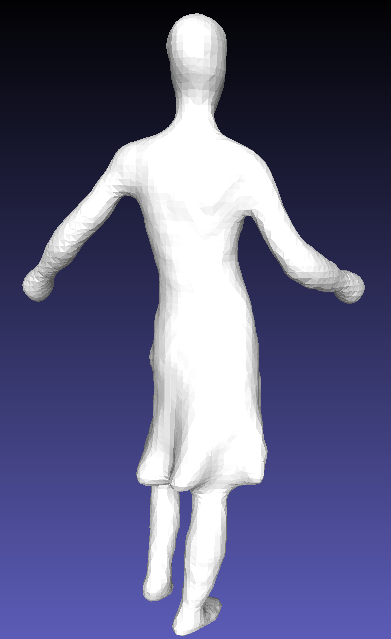}} &
\subfigure{\includegraphics[width=2cm, height=2.5cm]{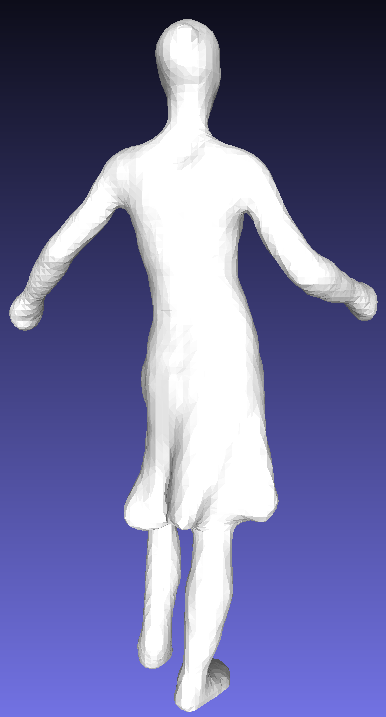}} &
\subfigure{\includegraphics[width=2cm, height=2.5cm]{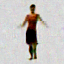}} &
\subfigure{\includegraphics[width=2cm, height=2.5cm]{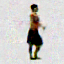}} &
\subfigure{\includegraphics[width=2cm, height=2.5cm]{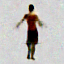}} &
\subfigure{\includegraphics[width=2cm, height=2.5cm]{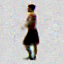}} &
\subfigure{\includegraphics[width=2cm, height=2.5cm]{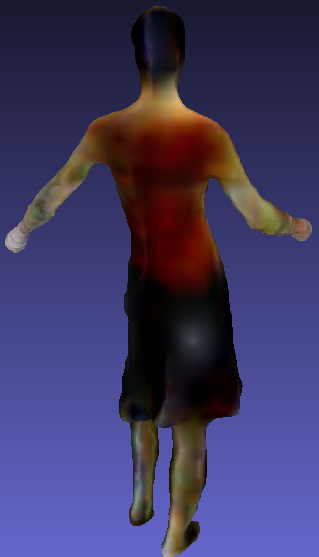}}  \\
\subfigure{\includegraphics[width=2cm, height=2.5cm]{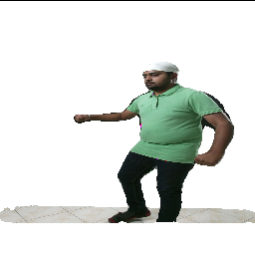}} &
\subfigure{\includegraphics[width=2cm, height=2.5cm]{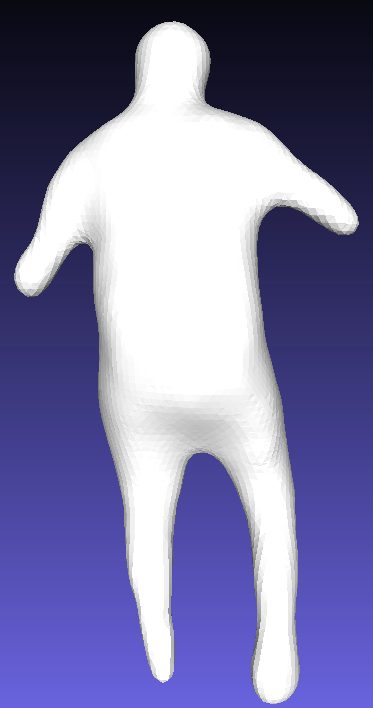}} &
\subfigure{\includegraphics[width=2cm, height=2.5cm]{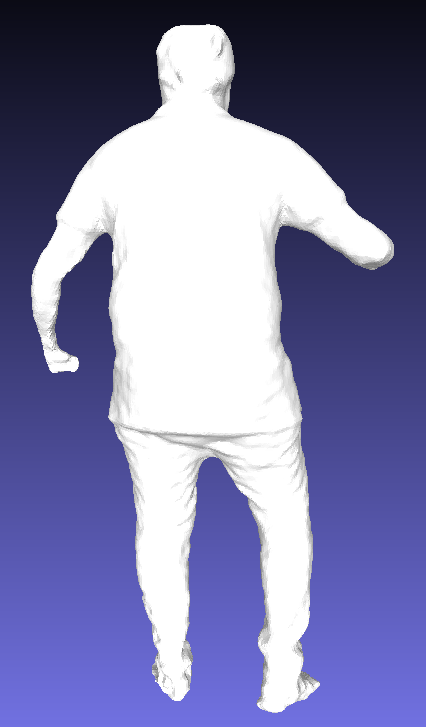}} &
\subfigure{\includegraphics[width=2cm, height=2.5cm]{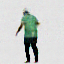}} &
\subfigure{\includegraphics[width=2cm, height=2.5cm]{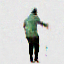}} &
\subfigure{\includegraphics[width=2cm, height=2.5cm]{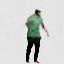}} &
\subfigure{\includegraphics[width=2cm, height=2.5cm]{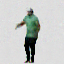}} &
\subfigure{\includegraphics[width=2cm, height=2.5cm]{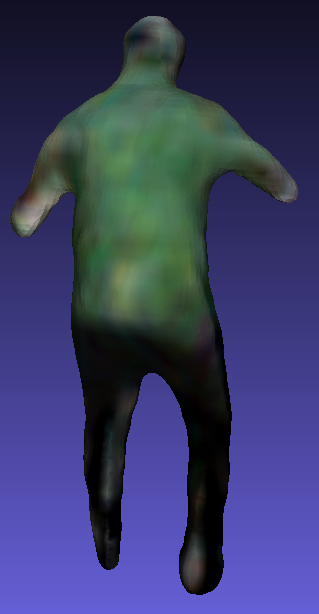}} \\
\end{tabular}
\end{adjustbox}
}
\caption{Results on reconstruction and texture recovery on ~\cite{VlasicMIT2008,SMPL2016}} and on our real data.
\label{fig:texture}
\end{figure*}
\section{Conclusion}
%Limitations.. extensions.. 
We proposed a novel deep learning pipeline for reconstructing textured 3D models of non-rigid human body shapes using single view RGB images (during test time). This is a severely ill posed problem due to self-occlusions caused by complex body poses and shapes, clothing obstructions, lack of surface texture, background clutter, single view, etc. %Further, a calibration-free environment adds additional complexity to both - reconstruction and texture recovery.  In this paper, we propose a deep learning based solution for textured 3D reconstruction of human body shapes from single view RGB input. 
%We attempted to overcome these challenges by first recovering the volumentric shape of non-rigid human body shapes given a single view RGB image by co-learning the RGB and depth cues at training time followed by orthographic texture view synthesis using the respective depth projection of reconstructed (volumetric) shape and input RGB image. 
We show superior textured reconstruction performance using the proposed method in terms of quantitative and qualitative results on both publicly available datasets (by simulating the depth channel with virtual Kinect) as well as real RGBD data collected with a calibrated multi Kinect setup. As a part of future work, it will be practical to extend this to recovering high resolution textured models and also exploiting the temporal consistency for the task of reconstruction, for the case of continuous mesh sequences. 
\newpage
%\bibliography{egbib}

\end{document}